\documentclass{article}
\usepackage{spconf,amsmath,graphicx,hyperref,booktabs,multirow,array}
\usepackage{fix-cm}
\usepackage{threeparttable}
\usepackage{balance}
\hypersetup{hidelinks}

\title{DyWPE: Signal-Aware Dynamic Wavelet Positional Encoding for Time Series Transformers}
\name{Habib Irani, Vangelis Metsis}
\address{Computer Science Department, Texas State University \\ San Marcos, TX 78666, USA}

\begin{document}
%
\maketitle
\begin{abstract}
Existing positional encoding methods in transformers are fundamentally signal-agnostic, deriving positional information solely from sequence indices while ignoring the underlying signal characteristics. This limitation is particularly problematic for time series analysis, where signals exhibit complex, non-stationary dynamics across multiple temporal scales. We introduce Dynamic Wavelet Positional Encoding (DyWPE), a novel signal-aware framework that generates positional embeddings directly from input time series using the Discrete Wavelet Transform (DWT). Comprehensive experiments on ten diverse time series datasets demonstrate that DyWPE consistently outperforms state-of-the-art positional encoding methods, with particularly significant improvements on longer sequences and complex biomedical signals.

\end{abstract}
\section{Introduction}
\label{sec:intro}

The transformer architecture has revolutionized sequential data modeling across diverse domains, from natural language processing to time series analysis \cite{vaswani2017attention}. A fundamental component enabling transformers to process sequential data is positional encoding, which addresses the inherent permutation invariance of self-attention mechanisms by injecting positional information into input representations.

In time series analysis, the importance of positional encoding is amplified due to the intrinsic temporal dependencies and complex multi-scale patterns characteristic of temporal data \cite{zerveas2021transformer, lim2021temporal}. However, existing positional encoding methods, ranging from sinusoidal encodings \cite{vaswani2017attention} to sophisticated relative positioning schemes \cite{shaw2018self, kerethinking}, share a fundamental limitation: they are signal-agnostic.
These methods derive positional information exclusively from abstract sequence indices (0, 1, ..., L-1) while remaining completely oblivious to the underlying signal characteristics. For instance, consider two time series segments occurring at identical absolute positions but exhibiting vastly different temporal dynamics: one representing a quiet, stable period with minimal variation, and another capturing volatile, high-frequency oscillations. Traditional positional encodings assign identical positional representations to both contexts, failing to capture the distinct temporal signatures that are crucial for effective time series modeling.
This signal-agnostic approach becomes particularly problematic when dealing with non-stationary signals, where statistical properties change over time, or multi-scale phenomena where different frequency components carry distinct semantic meanings. Recent comprehensive studies have highlighted significant performance variations across different positional encoding strategies in time series applications \cite{foumani2024improving}, yet no existing method addresses the fundamental limitation of signal-independent positioning.

To address this, we introduce Dynamic Wavelet Positional Encoding (DyWPE), a novel signal-aware framework that provides a more powerful inductive bias for temporal data. While a sufficiently large transformer can theoretically learn the complex relationships between signal-agnostic positions and signal content, this forces the model to learn the fundamental principles of time series dynamics from scratch. DyWPE makes this learning task more efficient by directly encoding the signal's local, multi-scale characteristics \textit{into} the positional representation itself. It achieves this by leveraging the Discrete Wavelet Transform (DWT) and learnable gating mechanisms to generate positional embeddings from the signal's content, creating a rich representation that dynamically adapts to local behavior. This approach offloads the burden of local pattern recognition, allowing the self-attention layers to focus more effectively on capturing long-range, higher-level dependencies.

Our key contributions are: (1) The first signal-aware positional encoding framework that derives positional information from signal content rather than sequence indices; (2) A computationally efficient implementation using DWT/IDWT operations with linear $\mathcal{O}(L)$ complexity; (3) Comprehensive experimental validation across ten diverse datasets demonstrating consistent superiority over eight established methods; (4) An ablation study examining the effectiveness and necessity of the different components of our algorithm, such as dynamic modulation and multi-scale wavelet decomposition.

\section{Background and Related Work}
\label{sec:related}

The application of attention in time series analysis has evolved from augmenting recurrent models to forming the core of modern transformers. Early work in sequence-to-sequence learning demonstrated the power of attention in encoder-decoder frameworks \cite{bahdanau2014neural}. For time series, this led to methods like dual-stage attention-based RNNs, which could selectively focus on relevant features and time steps for improved forecasting \cite{qin2017dual}. The success of these hybrids motivated the development of pure self-attention models, such as Gated Transformer Networks, which use learnable gates to capture discriminative temporal patterns without recurrence \cite{liu2021gated}.

The transformer architecture \cite{vaswani2017attention}, with its self-attention mechanism, offered a powerful new paradigm. Adaptations for time series were swift, with initial frameworks focusing on stable training for multivariate data \cite{zerveas2021transformer} and specialized multi-head attention mechanisms, as seen in the Temporal Fusion Transformer (TFT) \cite{lim2021temporal}. To better handle the unique structure of temporal data, recent models have also incorporated patch-based tokenization, which segments the time series to capture both local and global features \cite{cordonnier2021differentiable}.

A core challenge in these adaptations is the transformer's inherent permutation invariance, which necessitates an explicit Positional Encoding (PE) to inform the model of the temporal order. The original sinusoidal PE \cite{vaswani2017attention} provides a fixed, deterministic mapping based on sequence indices. To improve flexibility, subsequent research introduced learnable absolute embeddings and relative position representations \cite{shaw2018self}, which encode the distance between tokens. More advanced methods like Rotary Position Embedding (RoPE) \cite{su2024roformer} and Transformer with Untied Positional Encoding (TUPE) \cite{kerethinking} further refined these concepts by integrating positional information directly into the attention computation. This evolution reflects a broader trend towards more expressive and context-dependent positional representations, often at the cost of increased computational complexity.

Despite these advances, a recent survey of PE methods in time series transformers \cite{irani2025positional} highlights a shared, fundamental limitation: they are all \textbf{signal-agnostic}. These methods operate on integer indices and remain oblivious to the signal's content, treating a volatile period the same as a stable one. While specialized designs for time series have been proposed \cite{foumani2024improving, bell2023adaptive}, they still anchor their encodings to the position index. The most proximate work, by Oka et al. \cite{okawavelet}, proposes a wavelet-based PE for language models to improve length extrapolation. However, their method is also signal-agnostic, using wavelets to create a multi-scale representation of the \emph{relative distance between indices}, not the signal's content. To our knowledge, DyWPE is the first framework to break from this paradigm by constructing a positional encoding directly from the time series signal itself.

\section{Dynamic Wavelet Positional Encoding}
\label{sec:method}

\subsection{Problem Formulation and Overview}

Given a time series dataset $X = \{x_1, x_2, ..., x_n\}$ with $n$ samples, where each sample $x_i \in {R}^{L \times d_x}$ represents a $d_x$-dimensional time series of length $L$, and corresponding labels $Y = \{y_1, y_2, ..., y_n\}$ where $y_i \in \{1, 2, ..., c\}$, our objective is to learn a positional encoding that captures signal-specific temporal characteristics.

Traditional positional encodings follow $P = f(\text{indices})$ where $f$ is signal-independent. DyWPE introduces the paradigm $P = f(X, \theta)$, making positional encoding a learnable function of actual signal content through parameters $\theta$. This enables the model to distinguish between different temporal contexts (e.g., high-frequency transients vs. low-frequency trends) even at identical sequence positions.

\begin{figure}
    \centering
    \includegraphics[width=.95\columnwidth]{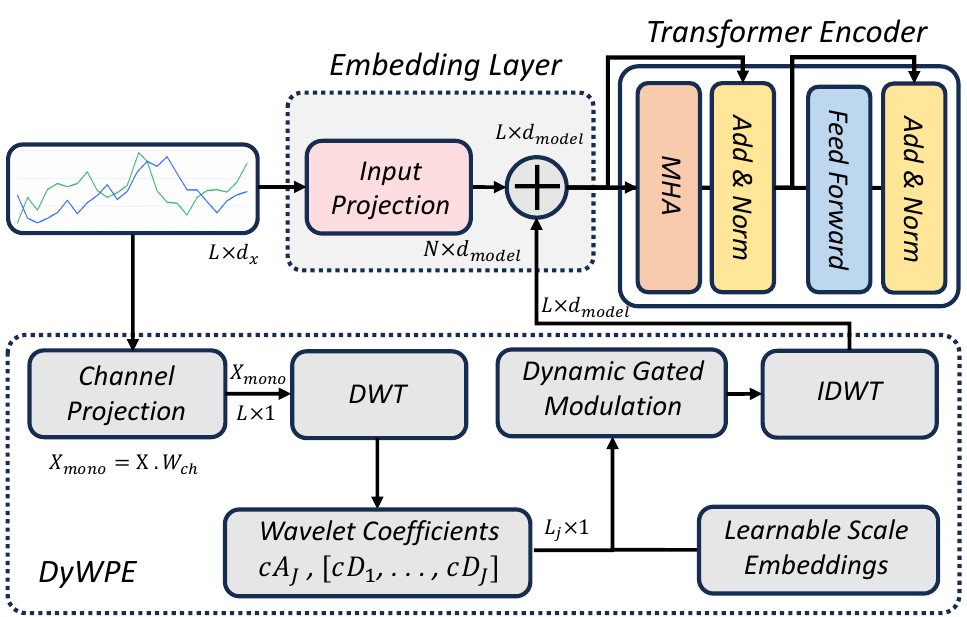}
    \caption{\small Transformer with DyWPE architecture overview showing the five-step process from multivariate input to final positional encoding through wavelet-based signal analysis.}
    \label{fig:dywpe_architecture}
    \vspace{-4mm}
\end{figure}

\subsection{Mathematical Formulation}

Given a multivariate time series $X \in {R}^{B \times L \times d_x}$ where $B$ is batch size, $L$ is sequence length, and $d_x$ is the number of input channels, DyWPE produces position embedding $P_{DyWPE} \in {R}^{B \times L \times d_{model}}$ through five sequential steps:

\textbf{Step 1: Channel Projection.} For multivariate signals, we create a single representative channel for wavelet analysis:
$$x_{mono} = x \cdot w_{channel}$$
where $w_{channel} \in {R}^{d_x}$ is a learnable projection vector capturing the most relevant temporal dynamics across input channels.

\textbf{Step 2: Multi-Level Wavelet Decomposition.} We apply $J$-level 1D Discrete Wavelet Transform to the projected signal:
$$(cA_J, [cD_J, cD_{J-1}, ..., cD_1]) = \text{DWT}(x_{mono})$$
This decomposition yields approximation coefficients $cA_J$ capturing low-frequency, large-scale trends and detail coefficients $cD_j$ for $j \in [1, J]$ capturing high-frequency, fine-scale patterns.

\textbf{Step 3: Learnable Scale Embeddings.} We introduce learnable embedding vectors serving as ``prototypes'' for each temporal scale:
$$E_{scales} = \{e_{A_J}, e_{D_J}, e_{D_{J-1}}, ..., e_{D_1}\}$$
where each embedding $e \in {R}^{d_{model}}$ corresponds to a specific scale captured by the DWT.

\textbf{Step 4: Dynamic Modulation.} The core innovation is the dynamic modulation mechanism, where actual wavelet coefficients modulate learnable scale embeddings through a gating function:
$$\text{gate}(e, c) = \left( \sigma(W_g e) \odot \tanh(W_v e) \right) \otimes c'$$
where $W_g, W_v \in {R}^{d_{model} \times d_{model}}$ are learnable weight matrices, $\sigma$ is sigmoid activation, and $c'$ is the coefficient tensor broadcasted appropriately.

This generates modulated coefficients for all scales:
\begin{align*}
Yl_{\text{mod}} &= \text{gate}(e_{A_J}, cA_J) \\
Yh_{\text{mod}} &= [\text{gate}(e_{D_J}, cD_J), ..., \text{gate}(e_{D_1}, cD_1)]
\end{align*}
\textbf{Step 5: Reconstruction.} The final positional encoding is reconstructed using Inverse DWT:
$$P_{DyWPE} = \text{IDWT}(Yl_{mod}, Yh_{mod})$$
leveraging the perfect reconstruction property of wavelets to synthesize modulated multi-scale information back into a sequence of length $L$.


\section{Experimental Evaluation}
\label{sec:experiments}

\subsection{Experimental Setup}

We conduct comprehensive experiments across ten diverse time series datasets spanning multiple domains such as Human Activity Recognition, Audio, and EEG classification, with eight datasets from the UEA archive \cite{bagnall2018uea} and two additional datasets \cite{micucci2017unimib, room_occupancy_estimation_864}, as shown in Table \ref{tab:datasets}.

\begin{table}[t]
\vspace{-5mm}
\caption{\small Time series dataset properties.}
\vspace{1.5mm}
\label{tab:datasets}
\fontsize{7.5}{9.3}\selectfont
\setlength{\tabcolsep}{4.5pt}
\renewcommand{\arraystretch}{1.1}
\begin{tabular}{@{}l c c c c c c@{}}
\toprule
\textbf{Dataset} & \textbf{Train} & \textbf{Test} & \textbf{Len} & \textbf{Cls} & \textbf{Ch} & \textbf{Type} \\
\midrule
Sleep (Sl) & 478,785 & 90,315 & 178 & 5 & 1 & EEG \\
ElectricDevices (ED) & 8,926 & 7,711 & 96 & 7 & 1 & Device \\
FaceDetection (FD) & 5,890 & 3,524 & 62 & 2 & 144 & EEG \\
MelbournePedestrian (MP) & 1,194 & 2,439 & 24 & 10 & 1 & Traffic \\
LSST (LS) & 2,459 & 2,466 & 36 & 14 & 6 & Other \\
SelfRegulationSCP1 (SR1) & 268 & 293 & 896 & 2 & 6 & EEG \\
SelfRegulationSCP2 (SR2) & 200 & 180 & 1152 & 2 & 6 & EEG \\
JapaneseVowels (JV) & 270 & 370 & 29 & 9 & 12 & AUDIO \\
UniMiB-SHAR (UM) & 4,601 & 1,524 & 151 & 9 & 3 & HAR \\
RoomOccupancy (RO) & 8,103 & 2,026 & 30 & 4 & 18 & Sensor \\
\bottomrule
\end{tabular}
\begin{flushleft}
\footnotesize
\end{flushleft}
\vspace{-6mm}
\end{table}

We evaluated DyWPE against eight established positional encoding methods using a patchTST. All experiments use consistent hyperparameters: 4 transformer layers, 4 attention heads, 128 hidden dimensions, dropout rate 0.2, and an Adam optimizer. The complete code implementation and benchmarks are made publicly available for reproducibility: {\small \url{https://github.com/imics-lab/DyWPE}}.

\subsection{Comparative Results}

\begin{figure}
    \centering
    \vspace{-4mm}
    \includegraphics[width=\columnwidth,clip]{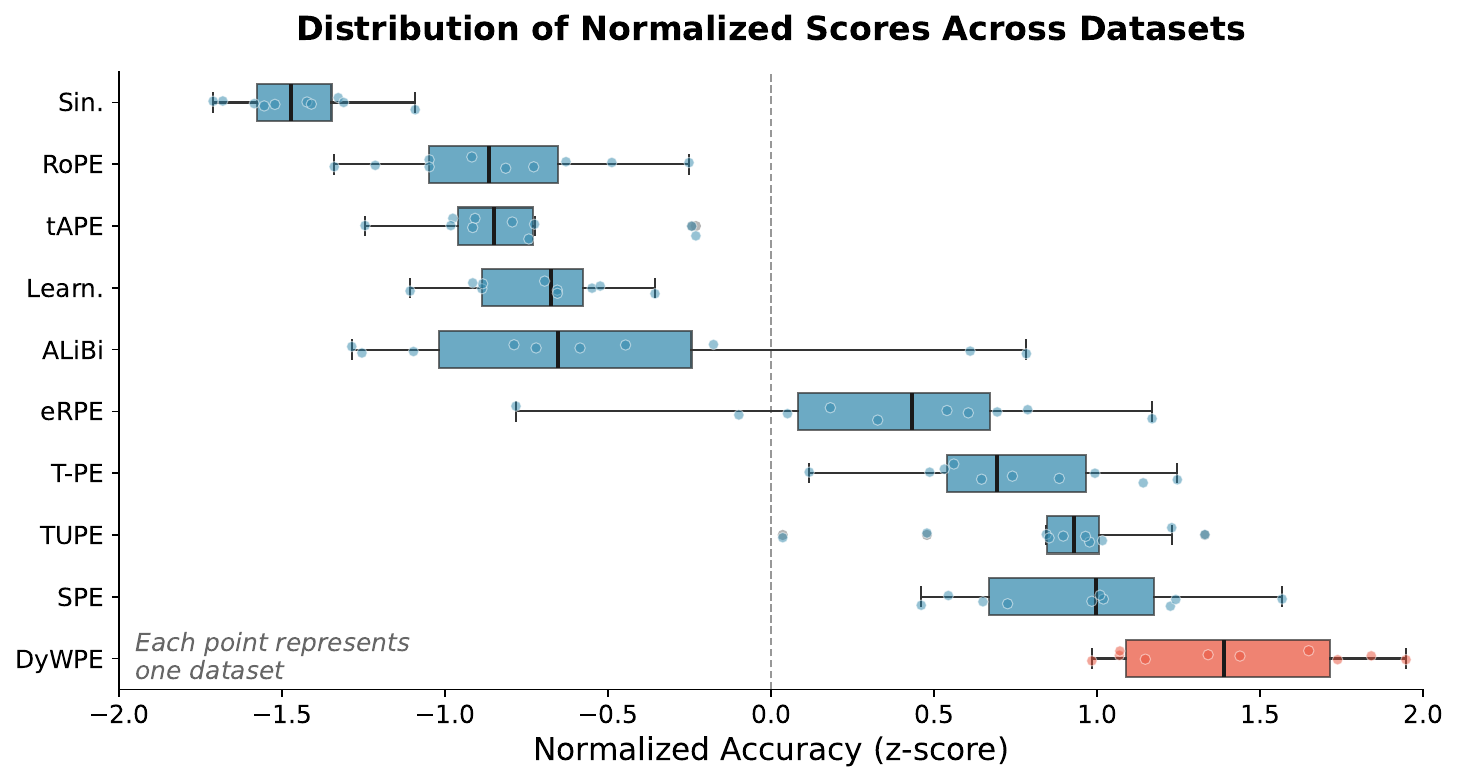}
    \vspace{-4mm}
    \caption{\small Distribution of z-score normalized (mean = 0, std = 1) classification accuracy across 10 datasets for each PE method. Box plots show the median (dark line), interquartile range (box), and data range (whiskers), with outliers shown as isolated points.}
    \label{fig:boxplot}
    \vspace{-4mm}
\end{figure}

Table~\ref{tab:main_results} presents comprehensive accuracy results across all datasets and methods, and Fig.~\ref{fig:boxplot} visualizes the same results in the form of a box plot. DyWPE demonstrates consistent superior performance, achieving the highest accuracy on 6 out of 10 datasets and ranking in the top 2 for the remaining datasets.

\begin{table}[t]
\centering
\caption{\small Classification accuracy comparison across all PE methods and datasets. All numbers are the average of 5 runs.}
\vspace{1.5mm}
\label{tab:main_results}
\fontsize{7.7}{8.9}\selectfont
\setlength{\tabcolsep}{2.5pt}
\renewcommand{\arraystretch}{1.05}
\begin{tabular}{l|cc|ccc|ccc|c}
\toprule
\multirow{2}{*}{\textbf{Data}} & \multicolumn{2}{c|}{\textbf{Absolute PE}} & \multicolumn{3}{c|}{\textbf{Relative PE}} & \multicolumn{3}{c|}{\textbf{Hybrid PE}} & \textbf{Ours} \\
& \textbf{Learn.} & \textbf{tAPE} &  \textbf{eRPE} & \textbf{ALiBi} & \textbf{SPE} & \textbf{RoPE} & \textbf{T-PE} & \textbf{TUPE} & \textbf{DyWPE} \\
\midrule
Sl & 85.2 & 85.1 & 86.4 & 85.3 & 87.6 & 84.4 & 87.2 & 87.9 & \textbf{88.2} \\
ED & 69.4 & 69.3 & 76.2 & 68.4 & \textbf{81.1} & 71.3 & 76.3 & 77.9 & 79.1 \\
FD & 64.2 & 65.3 & 67.8 & 62.6 & 67.4 & 63.5 & 68.6 & \textbf{69.5} & 68.8 \\
MP & 70.2 & 68.2 & 73.3 & 67.2 & \textbf{75.3} & 69.0 & 74.2 & 74.5 & 74.8 \\
LS & 58.2 & 58.4 & 61.1 & 59.2 & 60.1 & 58.3 & 60.2 & 59.5 & \textbf{62.2} \\
SR1 & 84.4 & 84.2 & 85.6 & 84.9 & 88.3 & 83.1 & 87.1 & 87.5 & \textbf{89.3} \\
SR2 & 54.6 & 53.3 & 56.4 & 58.6 & 58.2 & 53.1 & 56.6 & 59.3 & \textbf{61.2} \\
JV & 95.8 & 95.8 & 96.0 & 98.1 & 98.7 & 96.8 & 98.9 & 97.9 & \textbf{99.2} \\
UM & 84.4 & 83.3 & 86.4 & 84.1 & 86.7 & 83.6 & \textbf{87.1} & 86.5 & 86.7 \\
RO & 91.1 & 92.2 & 92.9 & 91.5 & 93.4 & 91.7 & 93.1 & 93.7 & \textbf{94.8} \\
\bottomrule
\end{tabular}
\vspace{-4mm}
\end{table}

\subsection{Performance Analysis}

\textbf{Sequence Length Effects}: DyWPE demonstrates strong performance across varying sequence lengths, with particularly notable advantages on longer, complex sequences. On the longest sequence (SR2, 1152 timesteps), DyWPE achieves 61.2\% accuracy, substantially outperforming most methods. For other long sequences, DyWPE shows consistent improvements: Sleep (88.2\%), SR1 (89.3\%), and competitive performance on ElectricDevices and UniMiB-SHAR. DyWPE maintains robust performance across diverse signal types and lengths, demonstrating the value of signal-aware positioning for complex temporal patterns.

\textbf{Domain-Specific Performance}: In biomedical signal processing (Sleep EEG, SelfRegulationSCP1, SelfRegulationSCP2), DyWPE demonstrates consistent improvements, achieving top performance on these datasets by effectively capturing multi-scale physiological dynamics. For sensor and device data, results are more variable: RoomOccupancy shows strong performance (94.8\%), while ElectricDevices lags behind specialized methods like SPE.

\begin{figure}
    \centering
    \includegraphics[width=0.95\columnwidth,clip]{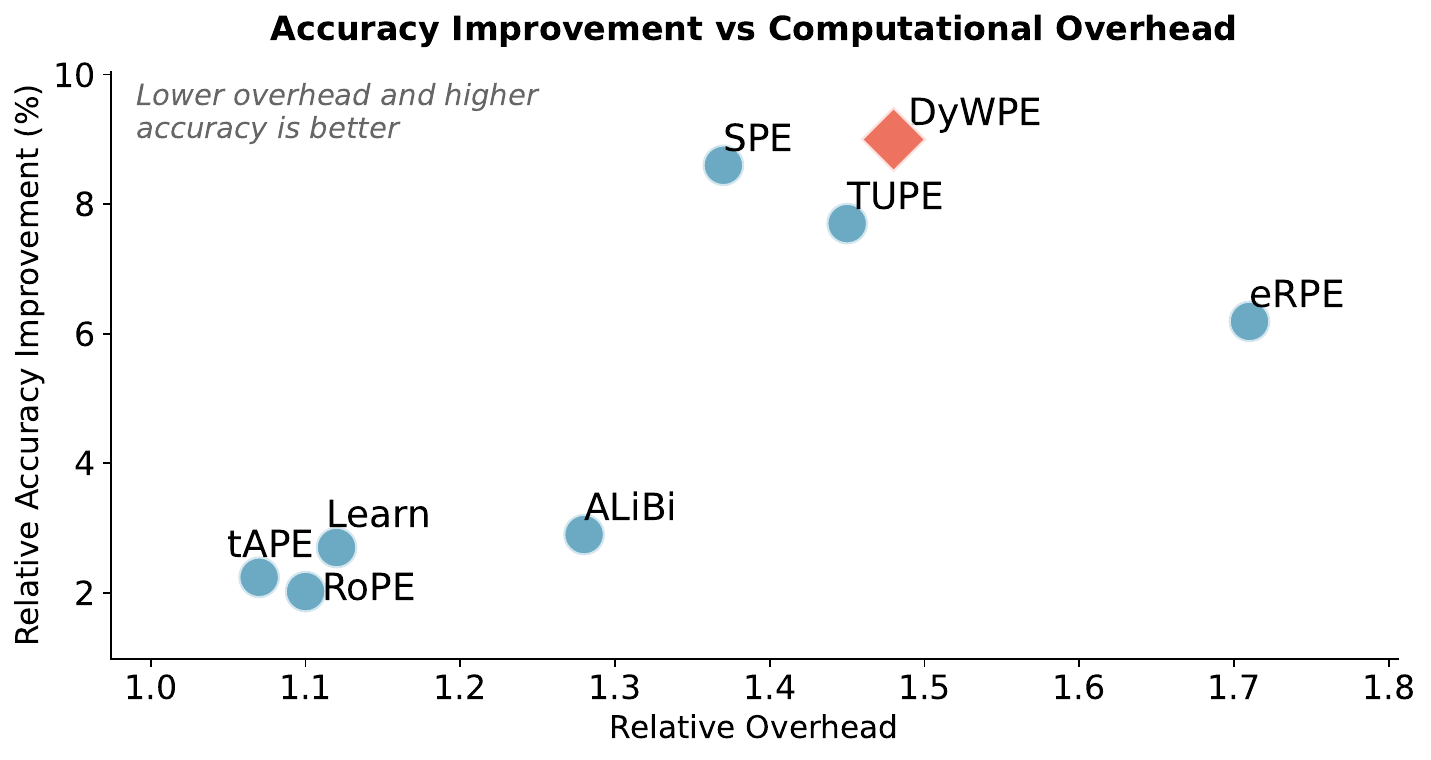}
    \vspace{-3mm}
    \caption{\small Average relative accuracy improvement vs. computational overhead trade-off of different SOTA PE methods on all datasets.}
    \label{fig:Performance}
    \vspace{-3mm}
\end{figure}

\textbf{Computational Efficiency}: The asymptotic time and space complexities of our method scale linearly with the length of the sequence, unlike other SOTA PE methods, which scale quadratically. The fact that our method uses the input signal adds some practical computational overhead compared to signal-agnostic methods; however, the relative added overhead does not exceed that of other SOTA methods. Table~\ref{tab:complexity} provides a detailed comparison of computational characteristics across different positional encoding methods, and Figure~\ref{fig:Performance} provides a visual depiction of the tradeoff between accuracy gains for different PE methods versus practical computational overhead compared to the baseline.

\begin{table}
\centering
\caption{\small Computational Complexity Comparison of Positional Encoding Methods and Training Time Analysis for PE methods on all Dataset. Parameters: $L$ = sequence length, $d$ = dimension, $h$ = attention heads, $l$ = transformer layers, $K$ = kernel size, $R$ = representation dimension. Relative Overhead is average on 10 datasets based on baseline model (No PE).}
\vspace{0.8mm}
\label{tab:complexity}
\resizebox{1\columnwidth}{!}{%
\fontsize{7.2}{8.5}\selectfont
\setlength{\tabcolsep}{3pt}
\renewcommand{\arraystretch}{1.1}
\begin{tabular}{l|cccc}
\toprule
\textbf{Method} & \textbf{Params} & \textbf{Memory} & \textbf{Time Complex.} & \textbf{Rel. Overh.} \\
\midrule
tAPE & $Ld$ & $\mathcal{O}(Ld)$ & $\mathcal{O}(Ld)$  & 1.07\\
RoPE & 0 & $\mathcal{O}(Ld)$ & $\mathcal{O}(L^2d)$  & 1.10\\
Learn. & $Ld$ & $\mathcal{O}(Ld)$ & $\mathcal{O}(Ld)$ & 1.12 \\
ALiBi & 0 & $\mathcal{O}(L^2h)$ & $\mathcal{O}(L^2h)$ & 1.28 \\
SPE & $3Kdh{+}dl$ & $\mathcal{O}(LKR)$ & $\mathcal{O}(LKR)$ & 1.37\\
TUPE & $2dl$ & $\mathcal{O}(Ld{+}d^2)$ & $\mathcal{O}(Ld{+}d^2)$ & 1.45\\
\textbf{DyWPE} & $2d^2 + \lfloor\log_2(L)\rfloor d$ & $\mathcal{O}(Ld)$ & $\mathcal{O}(Ld)$ & 1.48 \\
eRPE & $(L^2{+}L)l$ & $\mathcal{O}(L^2d)$ & $\mathcal{O}(L^2d)$  & 1.71\\
T-PE & $2d^2l/h{+}(2L{+}2l)d$ & $\mathcal{O}(L^2d)$ & $\mathcal{O}(L^2d)$ & 1.95 \\

\bottomrule
\end{tabular}
}
\end{table}

\subsection{Ablation Study}
\label{sec:ablation}



\begin{table}[t]
\centering
\caption{\small 
DyWPE: complete signal-aware multi-scale approach; Static PE: multi-scale framework with fixed coefficients; Single-Scale: signal-aware with J=1 decomposition.}
\vspace{1.5mm}
\label{tab:ablation}
\resizebox{\columnwidth}{!}{
\fontsize{7.5}{8.5}\selectfont
\setlength{\tabcolsep}{3pt}
\renewcommand{\arraystretch}{1.1}
\begin{tabular}{l|cccccccccc}
\toprule
\textbf{Method} & \textbf{Sl} & \textbf{ED} & \textbf{FD} & \textbf{MP} & \textbf{LS} & \textbf{SR1} & \textbf{SR2} & \textbf{JV} & \textbf{UM} & \textbf{RO} \\
\midrule
DyWPE & \textbf{88.2} & 79.1 & \textbf{68.8} & \textbf{74.8} & \textbf{62.2} & \textbf{89.3} & \textbf{59.4} & \textbf{99.2} & 86.7 & \textbf{94.8} \\
Static Wavelet & 86.5 & 77.4 & 67.4 & 74.1 & 61.0 & 87.7 & 57.9 & 98.4 & 85.8 & 93.8 \\
Single-Scale & 86.1 & 75.9 & 66.4 & 74.5 & 59.5 & \textbf{89.3} & 52.1 & \textbf{99.2} & \textbf{87.7} & 93.3 \\
\bottomrule
\end{tabular}
}
\vspace{-5mm}
\end{table}

We conduct three critical ablation experiments to validate DyWPE's core contributions: signal-awareness, multi-scale representation, and wavelet types.

\textbf{Signal-Awareness Validation:} To isolate the impact of signal-aware modulation, we compare DyWPE against Static Wavelet PE (SWPE), which retains the complete DWT/IDWT framework and learnable scale embeddings but removes signal dependency. SWPE uses fixed, learnable tensors instead of dynamic coefficient modulation, creating a signal-agnostic baseline with identical architectural complexity. Results in Table~\ref{tab:ablation} demonstrate that signal-awareness provides improvements across all 10 datasets (average +1.09\%), with particularly strong gains on Sleep (+1.7\%), SR2 (+1.5\%), and FaceDetection (+1.4\%). This confirms that dynamic adaptation to signal characteristics drives performance improvements beyond the multi-scale framework alone.

\textbf{Multi-Scale Analysis Validation:} We evaluate the necessity of hierarchical temporal decomposition by comparing full DyWPE against simplified single-scale variants. The multi-scale analysis shows variable effectiveness across datasets (7 out of 10 show improvements, average +1.60\%), with exceptional gains on SR2 (+7.3\%) and substantial improvements on Sleep (+2.1\%) and FaceDetection (+2.4\%). However, UniMiB-SHAR shows negative results (-1.0\%), indicating that multi-scale decomposition benefits depend on signal complexity. Datasets with rich temporal hierarchies benefit most from multi-scale analysis, while simpler patterns may not require deep decomposition.

\textbf{Effects of Different Wavelet Types:} We evaluated 11 wavelet families (Daubechies, Coiflets, Biorthogonal, Haar). Our experiments show that while db4 remains a robust default, bior2.2 showed a slight improvement on complex signals, suggesting that biorthogonal wavelets' reconstruction properties may further benefit signal-aware encoding.





\vspace{-2mm}
\section{Conclusion}
\label{sec:conclusion}
\vspace{-2mm}

We introduced Dynamic Wavelet Positional Encoding (DyWPE), the first signal-aware position encoding framework for time series classification. By analyzing actual signal content through multi-scale wavelet decomposition and dynamically modulating learnable scale embeddings, DyWPE creates rich positional representations that adapt to local temporal characteristics. Comprehensive experiments demonstrate consistent superiority over SOTA methods, with particularly significant improvements on longer sequences and complex signals.

\balance
\renewcommand{\refname}{References}
\bibliographystyle{plain}
{\fontsize{10}{10}\selectfont 
\bibliography{refs}
}

\end{document}